\documentclass[11pt]{asaproc}

\newcommand{\loss}{\mathcal L}		
\newcommand{\real}{{\rm I\!R}}
\newcommand{\eeta}{\boldsymbol{\eta}}
\newcommand{\ttheta}{\boldsymbol{\theta}}

\newcommand{\w}{\mathbf{w}}
\newcommand{\x}{\mathbf{x}}
\newcommand{\h}{\mathbf{h}}

\usepackage{amsmath}
\usepackage{natbib}
\usepackage{amsfonts}
\usepackage{graphicx}     
\usepackage{multicol}
\usepackage{multirow}
\usepackage{booktabs}

\newtheorem{theorem}{Theorem}

\usepackage{times}

\title{Activation Adaptation in Neural Networks}

\author{Farnoush Farhadi\thanks{Department of Mathematics and Industrial Engineering, Polytechnique Montreal, Montreal, Canada.} \and  Vahid Partovi Nia \thanks{Huawei Noah's Ark Lab, Montreal Research Centre, Montreal, Canada} \and  Andrea Lodi$^*$}
\begin{document}

\maketitle

\begin{abstract}
Many neural network architectures rely on the choice of the activation function for each hidden layer. Given the activation function, the neural network is trained over the bias and the weight parameters. The bias catches the center of the activation, and the weights capture the scale. Here we propose to train the network over a shape parameter as well. This view allows each neuron to tune  its own activation function and adapt the neuron curvature towards a better prediction. This modification only adds one further equation to the back-propagation for each neuron. 
Re-formalizing activation functions as CDF generalizes the class of activation function extensively. We aimed at generalizing an extensive class of activation functions to study: i) skewness and ii) smoothness of activation functions. 
Here we introduce adaptive Gumbel activation function as a bridge between Gumbel and sigmoid. A similar approach is used to invent a smooth version of ReLU. Our comparison with common activation functions suggests different data representation especially in early neural network layers. This adaptation also provides prediction improvement.
\end{abstract}


\section{Introduction}
Neural networks achieved considerable success in image, speech, and text classification. In many neural networks only bias and weight parameters are learned to fit the data, while the activation function of each neuron is pre-specified to sigmoid, hyperbolic tangent, ReLU, etc. From a theoretical standpoint, a neural network reasonably wide and deep, approximates an arbitrarily complex function  independent of the chosen activation function \citep[][]{hornik1989multilayer, cho2010large}. However, in practice, the prediction performance and the learned representation depends on  hyperparameters such as network architecture, number of layers, regularization function, batch size, initialization, activation function, etc.  Despite large studies on network hyperparameter tuning, there have been few studies on how to choose an appropriate activation function. The choice of activation function changes learning representation and also affects the network performance \citep[][]{agostinelli2014learning}. We propose let data estimate the activation function during training by developing a flexible activation function. We demonstrate how to formalize this such activations and show how to embed it in the back-propagation.

Developing an adaptive activation function helps fast training of deep neural networks, and has attracted attention, see \cite[][]{zhang2015parameterised, agostinelli2014learning, jarrett2009best, glorot2011deep, goodfellow2013maxout, springenberg2013improving} and recently \cite{dushkoff2016adaptive,hou2016neural,hou2017convnets}. Here, we introduce adaptive activation functions by combining two main tools: i) looking at activation as a cumulative distribution function and ii) making an adaptive version by equipping a distribution  with a shape parameter. The shape parameter is continuous, so that an update equation can be added in back-propagation. Here, we focus on the simple architecture of LeNet5, but this idea can be used to equip more complex and deep architectures with flexible activations \citep{ramachandran2018searching}.

There has been a surge of work in modifying the ReLU. Leaky ReLU is one of the most famous modifications that gives a slight negative slope on a negative argument  \citep[][]{maas2013rectifier}. Another modification called ELU \citep[][]{clevert2015fast} attempts exponential decrease of the slope from a predefined value to zero, see laso GELUs of \cite{hendrycks2016gaussian}. Taking a mixture approach,  \citep[][]{qian2018adaptive} proposed a mixed function of leaky ReLU and ELU as an adaptive function, that could be learned in a data-driven way.   Inspired by \cite[][]{agostinelli2014learning, zhang2015parameterised} and \cite[][]{qian2018adaptive}, we study the effect of i) the asymmetry and ii) the smoothness of activation function.  The first study is performed by  introducing an adaptive asymmetric Gumbel activation that changes its shape towards the symmetric sigmoid function. The second study is achieved by equipping the ReLU function with a smoothness parameter. In both cases, we tune the shape parameter for each neuron independently by adding an updating equation to back-propagation.

The performance of two fully-connected neural networks and a convolutional network are compared on simulated data, MNIST benchmark, and Movie review sentiment data.  As an application, we use the classical LeNet5 architecture \citep[][]{kim2014convolutional} to classify the users' intention using URLs they navigated on their browser. 
 
\section{Adaptive Activations}
We recommend to perceive the activation function as a cumulative distribution function bounded on $[0,1]$. Common activation functions are bounded, but not necessarily to $[0,1]$ like hyperbolic tangent. A location and scale transformation is sufficient to transform their range if necessary. However, still widely-used  activations such as ReLU or leaky ReLU are unbounded. One may decompose unbounded activation functions into two components: i) a bounded component and ii) an unbounded component, and only adapt the bounded ingredient through a continuous cumulative distribution function.

\subsection{Adaptive Gumbel}
One of the common activation functions is the sigmoid function that maps a real value to $[0,1]$ similar to cumulative distribution functions. More precisely, the sigmoid function is the cumulative distribution function of the symmetric and bell-shaped logistic distribution. An asymmetric activation can be developed using a cumulative distribution function of a skewed distribution like Gumbel, for example. Gumbel distribution is  the asymptotic distribution of extreme values such as minimum or maximum. A shape parameter that pushes a sigmoid distribution function away from the logistic and more towards a Gumbel distribution defines a sort of a shape parameter.
Our proposed adaptive Gumbel activation is
\begin{equation}
	\sigma_\alpha(x) = 1-{\{1+\alpha\exp(x)\}^{-\frac{1}{\alpha}}} \quad \alpha\in\real^+, x\in \real.              
\label{eq:adagumbel}
\end{equation}

\begin{figure}[t]
  \centering
  \includegraphics[width=0.45\textwidth]{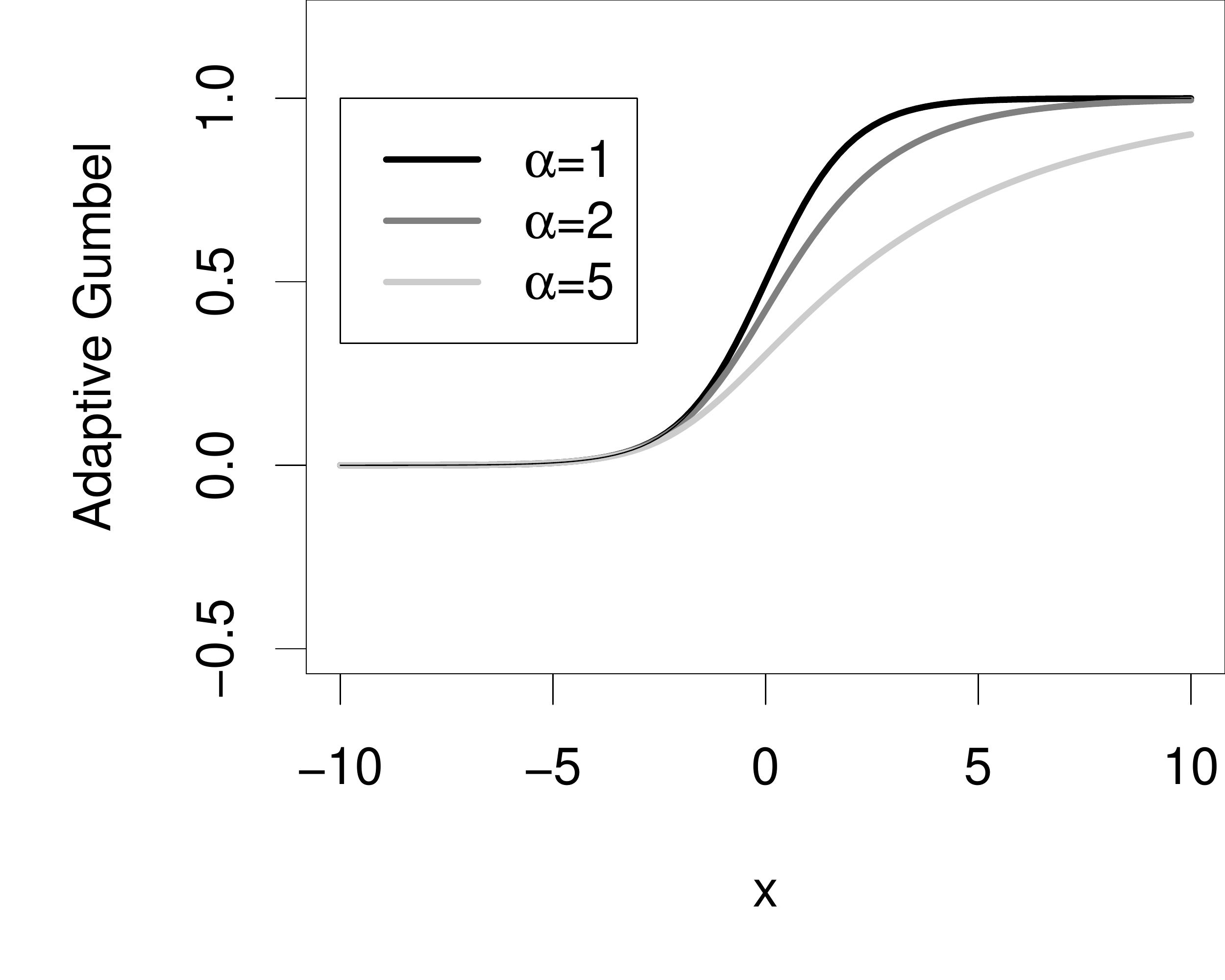}  \includegraphics[width=0.45\textwidth]{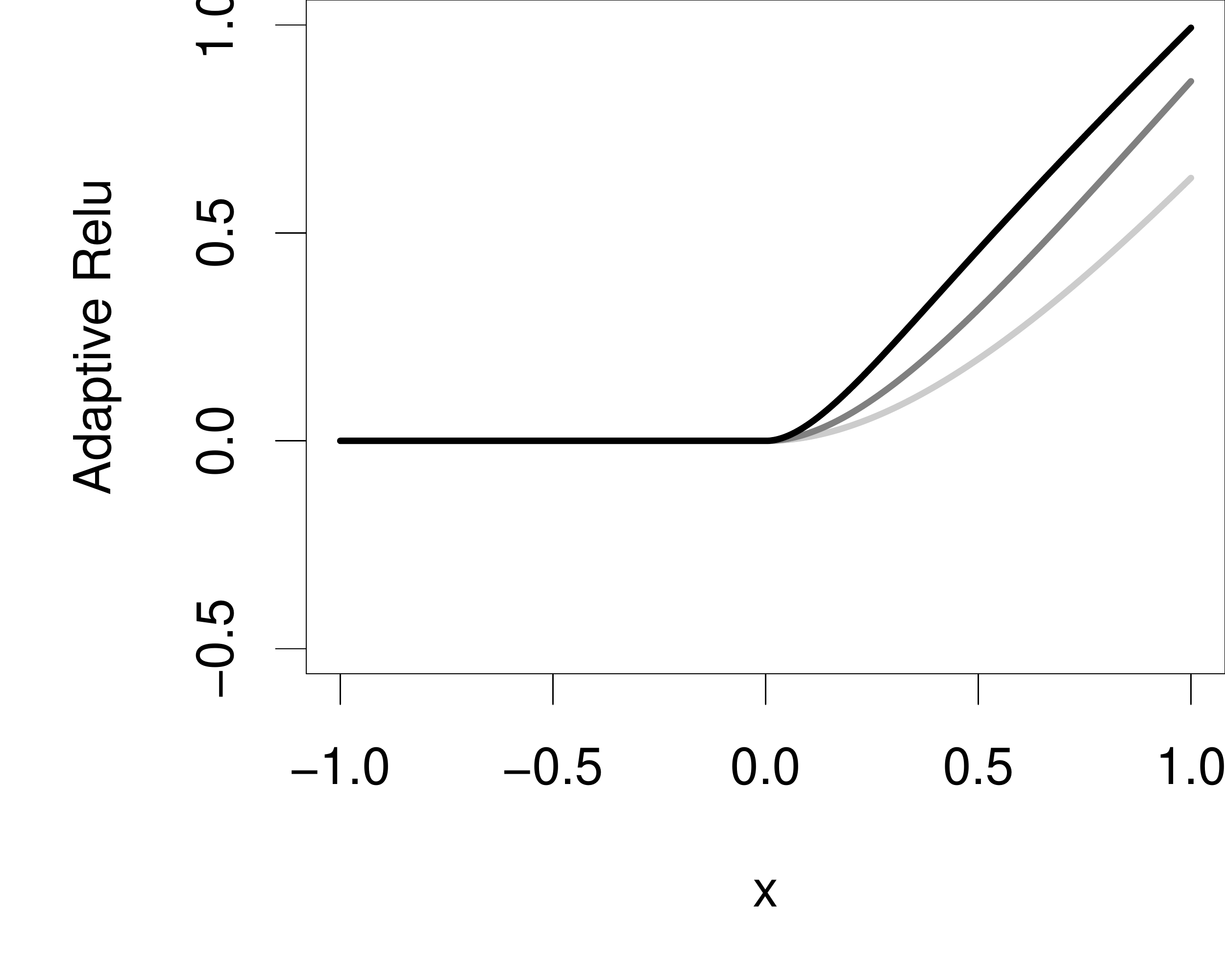} 
  \caption{Adaptive Gumbel activation (left panel) and adaptive ReLu activation (right panel) for $\alpha = 1, 2, 5$.}
  \label{fig:3_adaptive_sigmoid}
\end{figure}

The above form is inspired by Box-Cox transformation \citep{box1964analysis} for binary regression. The simplest form of neural network with no hidden layer is a binary regression in which  \eqref{eq:adagumbel} generalizes logistic regression towards complementary log-log regression by tuning $\alpha \in (0,1]$, see Figure~\ref{fig:3_adaptive_sigmoid} (left panel).  The Gumbel cumulative distribution function arises in the limit while  $\alpha\to 0$.

\subsection{Adaptive ReLU}
The ReLU activation function  
$$\sigma (x) = \max(0,x)$$ 
is unbounded, unlike the sigmoid or hyperbolic tangent. One may re-write the ReLU activation as 
$$\sigma (x) = x\Delta(x),$$ 
where $\Delta(x)$ is the cumulative distribution function of a degenerate distribution. The function $\Delta(x)$ is also known as \emph{Heaviside} function and coincides with the integral of the \emph{Dirac delta} function. We propose to replace  $\Delta(x)$ with a smooth cumulative distribution function $\Delta_\alpha(x)$ such as the exponential cumulative distribution function 
\begin{eqnarray}
\Delta_\alpha(x) &=& (1-e^{-\alpha x})\mathbb{I}_{\{x>0\}}(x),\nonumber\\ 
\sigma_\alpha(x) &=& x \Delta_\alpha(x), \quad \alpha\in \real^+, x\in \real, \label{eq:adaReLU}
\end{eqnarray}
where $\mathbb I_A(x)$ is the indicator function on set $A$.

In \eqref{eq:adaReLU} we recommend to equip the degenerate distribution $\Delta(x)$ with a smoothing parameter $\alpha$.  Any continuous random variable with a scale parameter is a convenient choice for $\Delta_\alpha(x)$. A random variable with infinitesimal scale behaves like a degenerate distribution, so $\Delta(x) $ is retrieved when the scale tends to zero, or equivalently $\alpha \to \infty$. The  generalized ReLU \eqref{eq:adaReLU} coincides with the SWISH activation function \citep{ramachandran2018searching} if $\Delta_\alpha(x)$ is the logistic cumulative distribution function
\begin{equation}
\Delta_\alpha(x) = (1+e^{-\alpha x})^{-1}.
\label{eq:swish}
\end{equation}
The SiLU \citep{elfwing2018sigmoid} is a special case of \eqref{eq:swish} while $\alpha=1$.

One may show that the proposed parameterization preserved identifiability if simple Bernoulli regression model.
\begin{theorem}
\label{theo:iden}
\label{LeftCosetsDisjoint}
a binary regression with the adaptive activation \eqref{eq:adagumbel} is identifiable. 
\end{theorem}
See Appendix for the proof.

\section{Back-propagation}
Define the vector of linear predictors of layer $l$ as 
$$\eeta^l=[\eta_1^l,\ldots,\eta^l_n]^\top,$$ 
where 
$$\eta^0_i = w_0 + \w^\top \x_i, i=1,\ldots,n,$$ 
and the $l$th hidden layer output $\h^l = \sigma(\eeta^l),$ in which $\eeta^l = w_0 + \w^\top \h^{l-1}$. 

Traditionally, $\sigma(x)$ is sigmoid or ReLU activation. The adapted back-propagation uses the conventional back-propagation, but each neuron carries its own activation function $\sigma_\alpha(x)$. The adaptation parameter $\alpha$ for each neuron is trained along with bias and  weights $[w_0, \w]^\top.$

Suppose the vector of  parameters for a neuron in layer $l$ is  $\ttheta^l = [\alpha, w_0, \w]$ and the network is trained using loss function $\loss(.)$.

In practice, $\loss(.)$ is the entropy loss for classification, and the squared error loss for regression, penalized with an $L_2$ norm $\sum_j \theta_j^2$ or an $L_1$ norm $\sum_j |\theta_j|$ upon convenience.  The updating back-propagation rule, given a learning rate $\gamma>0,$ for a neuron in layer $l$ is 
\begin{eqnarray}
w_0^l & \leftarrow & w_0^l -\gamma \frac{\partial \loss}{\partial w_0^l},\label{eq:BPbias}\\
\w^l  & \leftarrow &  \w^l - \gamma \frac{\partial \loss}{\partial \w^l},\label{eq:BPweight}\\
\alpha^l & \leftarrow & \alpha^l - \gamma \frac{\partial \loss}{\partial \alpha^l}, 
\label{eq:BPshape}
\end{eqnarray}
where \eqref{eq:BPbias} updates the bias, \eqref{eq:BPweight} updates the weights, and \eqref{eq:BPshape} adapts the activation function. Note that one may choose different learning rates for each equation. We recommend to reparametrize \eqref{eq:BPshape} with $e^\alpha$ in numerical computations to enforce $\alpha>0$.

\begin{figure}[t]
  \centering
  \includegraphics[width=.6\textwidth]{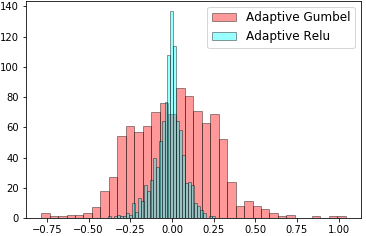}
  \caption{Histogram of fitted $\alpha$ for Adaptive ReLu (blue) and Adaptive Gumbel (red) activation functions for simulated data with one hidden layer.}
  \label{fig:4_parameter_training_hist}
\end{figure}

\begin{figure}[t]
  \centering
  \includegraphics[width=0.95\textwidth]{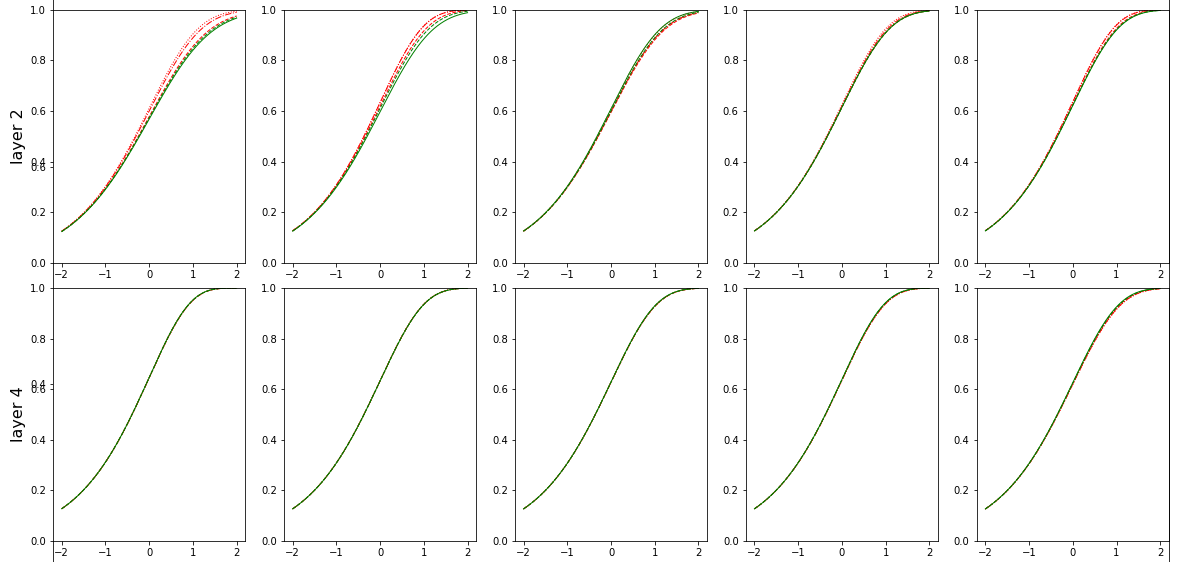}
  \caption{Fitted adaptive Gumbel activations on layer 2 and layer 4 of an eight hidden layer network. Adapted activations vary more often in earlier layers, see also Figure~\ref{fig:adarelu_8nn_activations}.}
  \label{fig:adagumbel_8nn_activations}
\end{figure}

\begin{figure}[t]
  \centering
  \includegraphics[width=.95\textwidth]{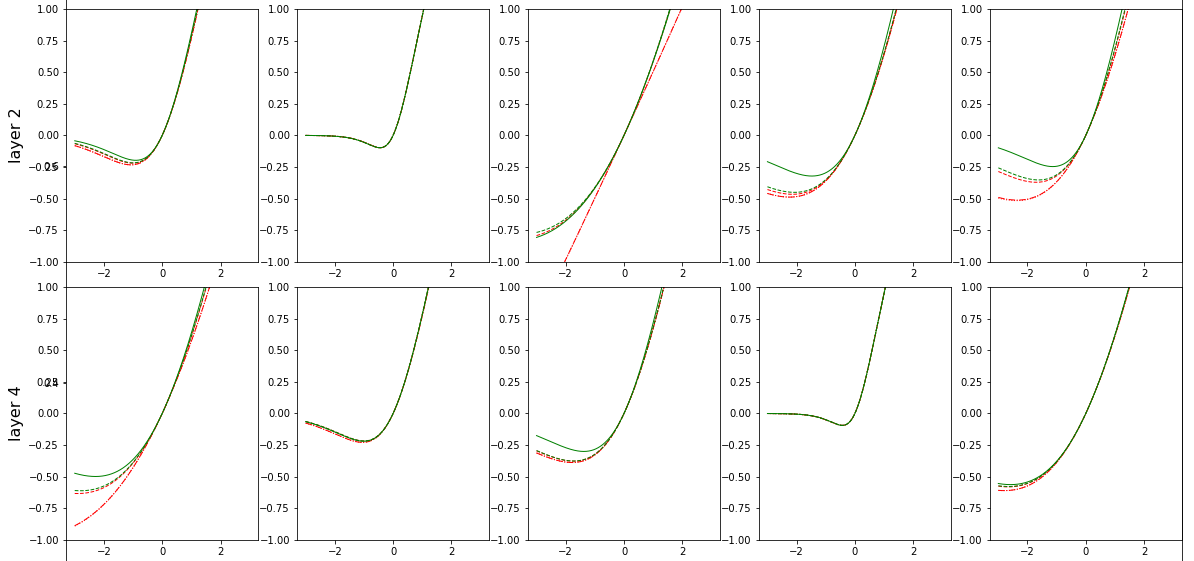}
  \caption{Fitted adaptive ReLU activations on layer 2 and layer 4 of an eight hidden layer network. Adapted activations vary more often in earlier layers, see also Figure~\ref{fig:adagumbel_8nn_activations}.}
  \label{fig:adarelu_8nn_activations}
\end{figure}

\section{Benchmarks}
Here, we compare the adaptive modification in three datasets. One dataset is a simulated fully-connected network in Section~\ref{sect:sim}, where the true activation function and true labels are  known. Furthermore, we evaluate activation adaptation  on convolutional models on the MNIST image data in Section~\ref{sect:mnist}, and on Movie Review text data in Section~\ref{sect:movie}. 

\subsection{Simulated Data}
\label{sect:sim}
Our objective in this experiment is to understand  how the choice of activation function  affects the performance of the network. We simulated data from two fully-connected neural networks: i) a network with only 1 hidden layer, and ii) with 8 hidden layers, each layer with 10 neurons. Activation functions are fixed to ReLU and  sigmoid in data simulation setup.  
"According to works in \citep[][]{lecun1998efficient} and recently in \citep[][]{krizhevsky2012imagenet} and \citep[][]{glorot2010understanding}, several weight initialization and different combination of number of input and output units in weight initialization formulation along with different type of activation functions could be employed in deep neural networks. In this paper, biases were initialized from $\mathcal N(0, 0.5)$ in simulated models and biases in fitted models are initialized
by formulation suggested in \citep[][]{lecun1998efficient}.
The origin weights in simulated models were initialized from a mixture of two normals $\mathcal{N}_1(1, 0.5)$ and  $\mathcal{N}_2(-1, 0.5)$ with an equal proportion and weights in fitted models initialized by \citep[][]{lecun1998efficient} settings."


"The simulated data set includes $10000$ examples of $10$ features with a binary output. In each configuration, we have a fully-connected network with 10 neurons at each layer is trained with a fixed learning rate $\gamma = .01$, regularization parameters $L_1 =.001$ and $L_2 = .001$, batch size = $20$, and number of epochs = $2000$ for both simulated and fitted models. The learning rate, $L_1$ and $L_2$ regularization constants are tuned using $5$-fold cross-validation. The average results are reported from a 5-fold cross validation as well."

The results summarized in Table~\ref{table:simu_accuracy} show that, overall, adaptive Gumbel outperforms sigmoid. Adaptive ReLU competes closely with ReLU in shallow networks, and slightly outperforms ReLU in deeper networks. Figure~\ref{fig:4_parameter_training_hist} confirms adaptive Gumbel and adaptive ReLU have different training range for $\alpha$. Figure~\ref{fig:adagumbel_8nn_activations} and Figure~\ref{fig:adarelu_8nn_activations} depicts the learned activations in an eight hidden layer fully connected network. Early layers have more variable learned activations, and often the last layers do not change much. The deeper layers are more difficult to learn. 

"It is worth mentioning that we run several models with different number of fully connected hidden layers from 1 to 8 including 1, 2, 4 and 8 layers. In Table 1, we aimed at studying the effect of varying activation functions on the accuracy of the predictions produced by the fitted model compared to its original counterpart when they have the same number of hidden layers, same number of neurons at each layer. Our goal was to understand how accurately fitted model could predict the class labels at the end of training step while we increase the number of hidden layers from 1 (as a simple fully connected multi-layer perceptron) to a very deeper one with 8 hidden layers. In this paper we only present the results for 1 and 8 hidden layers due to page limits. For models with less layers, fitting models with ReLu or adaptive ReLu almost outperforms the other activation functions. For deeper models, fitting models with the adaptive Gumbel also exhibits a good performance in competition with ReLu and adaptive ReLu. As seen in this table, as the number of hidden layers increases, an expected drop in the performance of all fitted models is observed. Hence, it is evident that there is no question to continue for deeper fully-connected layers."

\begin{table}
\centering
\begin{tabular}{|c c | c c  c c|}
\hline 
  \multicolumn{2}{|c|}{simulated}   & \multicolumn{4}{|c|}{fitted network}\\
&&&&&\\
layers  &  & Sig & AGumb & ReLU & AReLU\\
 \hline
\multirow{2}{*}{1} & Sig & $95.8$ & {97.5} & \textbf{97.8} & 97.7 \\
                     & ReLU & 96.1 &{97.4} & \textbf{98.2} & 98\\
\hline
\multirow{2}{*}{8}& Sig & 83.7 & \textbf{83.8} & 81.8 & {81.9}\\
   					& ReLU &  57.3 & {88.2} & 89.3 & \textbf{89.9}\\
\bottomrule[1.5pt]
\end{tabular}
\caption{Prediction accuracy for data simulated with sigmoid (Sig), adaptive Gumbel (AGumb), ReLU and adaptive ReLU (AReLU) in networks with 1 and 8 hidden layers. The maximum standard error is 0.22.}
\label{table:simu_accuracy}
\end{table}

\subsection{MNIST Data}
\label{sect:mnist}


Here we evaluate the performance of adapting activation on convolutional neural networks using handwritten digits grayscale image data. Our convolutional architecture is the classical LeNet5 \citep[][]{lecun1998gradient}, but with adaptive activations.  This architecture contains two convolutional layers, each convolutional layer followed by a max-pooling layer. A single fully-connected hidden layer is put on top with $1000$ neurons.

Motivated from Section~\ref{sect:sim}, we only keep ReLU as the strong competitor,  because adaptive Gumbel always outperforms sigmoid. The network parameters  are trained with batch normalization, learning rate $\gamma = 0.01$, batch size  $100$, and iterated $150$ epochs. Figure~\ref{fig:traincost} (left panel) shows are adaptive models converge.
Prediction accuracy is summarized in Table~\ref{tab:mnist_cnn}. The best performance appears for adaptive Gumbel on convolutional layer, closely followed by ReLU. 
"Hyper parameters including learning rate for CNN models are chosen according to the primary settings of LeNet5 implementation developed by Theano development team. We select fixed hyper parameters in all CNN models to study the effect of changing activation functions on final performance. The accuracy of each model is reported by running the corresponding algorithm on standard test set provided in MNIST data."

\begin{table}[ht!]
\centering
\begin{tabular}{|c l| c c c |}
\hline 
\multicolumn{2}{|c}{Conv layer} &\multicolumn{3}{|c|}{fully-connect layer}\\
	 & & ReLU  & AReLU & AGumb \\
\hline
\multirow{3}{*}{} 
& ReLU 	   &  {99.1} & 98.7 & \textbf{99.1}\\
& AReLU    &  98.8 &  {98.9}& {98.9}\\
& AGumb &  98.7 &   98.8& {98.9}\\
\hline
\end{tabular}
\caption{Convolutional architecture of LeNet5 on MNIST data. The activations in the rows represent the activation functions of convolutional layers, while the columns represent activations of fully-connected layers for ReLU, adaptive ReLU (AReLU), and adaptive Gumbel (AGumb) activations. The sigmoid activation is not reported as it was beaten always by the other techniques. The accuracy is computed over the standard test set.}
\label{tab:mnist_cnn}
\end{table}



\begin{figure}[t]
  \centering
  \includegraphics[width=0.45\textwidth]{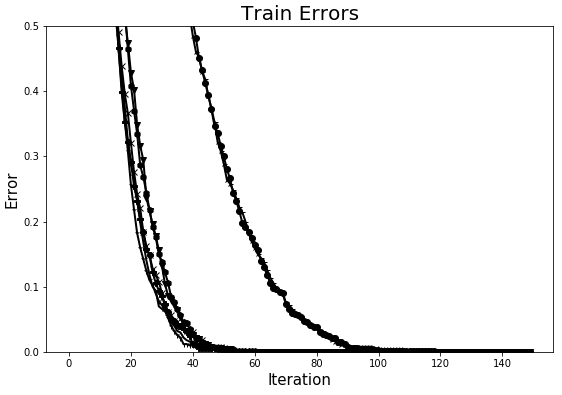}
  \includegraphics[width=0.45\textwidth]{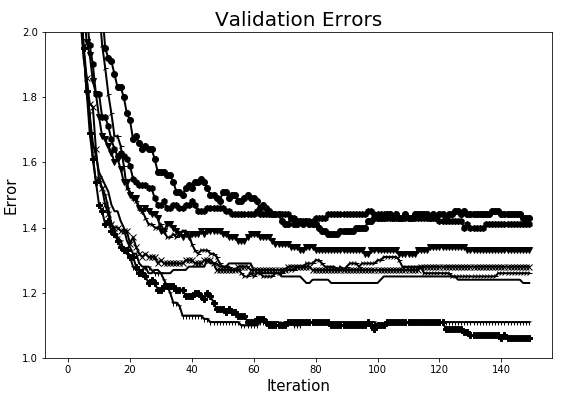}
  \caption{Training and validation error curves of different adaptive models on MNIST data.}
  \label{fig:traincost}
\end{figure}

\subsection{Movie Review Data}
This time we try convolutional architecture on text data over pre-trained word vectors. The data consists of 2000 movie reviews, 1000 positive and 1000 negative  \citep[][]{pang2004sentimental}. 

These word vectors use the word2vec  \citep[][]{mikolov2013distributed} trained on 100 billion words of Google News to embed a word in a  vector of dimension 300. Word2vec transforms each word into a vector such that the words semantics is preserved. 
"A CNN model with pre-trained word2vec vectors called static in \citep[][]{kim2014convolutional}, is used in our experiments. In this variant, the static CNN we use involves two convolutional layers each of them followed by a max-pooling layer and a fully-connected layer at the end. The fully-connected network includes one hidden layer with 100 neurons and a softmax output layer for binary text classification. The hyper parameters in all models are same including learning rate $\gamma = 0.05$, image dimensions (img-w) $= 300$, filter sizes $= [3, 4, 5]$, each have 100 feature maps, batch size $= 50$, dropout $= 0.5$, number of hidden layers $= 1$, number of neuron $= 100$ and number of epochs $= 50$.
For consistency, same data, pre-processing and hyper-parameter settings are used as reported in \citep[][]{kim2014convolutional}. Unlike MNIST data, the Movie Review dataset does not have a standard test set. So, we report the average accuracy over 5-fold cross-validation in Table~\ref{tab:movie_cnn}."

 Again adaptive activation provides a better prediction accuracy. Figure~\ref{fig:traincost} (right panel) suggests that for Movie review data is more difficult to converge compared to MNIST. We suspect this happens, because  i) text data carry less information compared to image, ii) embedding may mas some information that exists in the text.

\begin{figure}[ht]
  \centering
  \includegraphics[width=.95\textwidth]{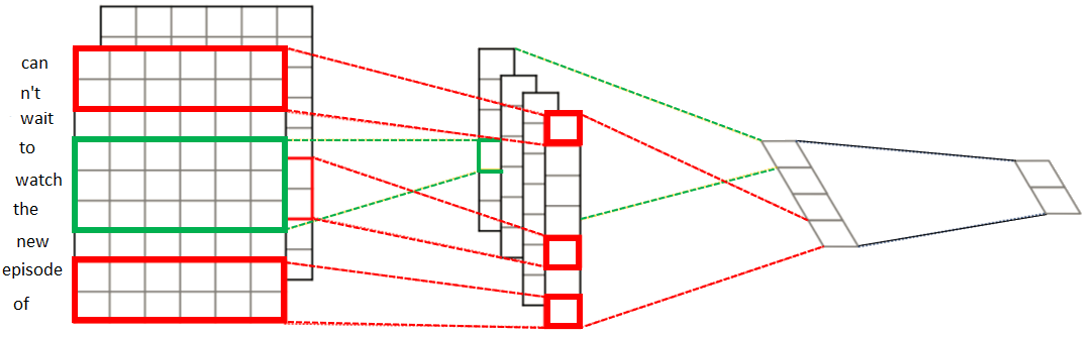}
  \caption{Movie comments are embedded into a vector. Then a CNN model classifies the text to a "positive review" or "negative review".}
  \label{fig:moviearch}
\end{figure}

\begin{table}[ht!]
\centering
\begin{tabular}{|l l| c c c c|}
\hline
\multicolumn{2}{|c|}{Conv layer} &\multicolumn{4}{c|}{Fully-connect layer}\\
	 & & & ReLU & AReLU & AGumb\\
\hline
 & ReLU& & {79.1} & {79.1} & 78.9\\
& AReLU&   & 78.9 & 78.5 & {\bf 79.3}  \\
& AGumbel  &   & 52.3 & 77.4  & {78.7}  \\
\hline
\end{tabular}
\caption{Convolutional architecture of LeNet5 on Movie review  data. The activations in the rows represent the activation functions of convolutional layers, while the columns represent activations of fully-connected layers for ReLU, adaptive ReLU (AReLU), and adaptive (AGumb) Gumbel activations. The maximum standard error estimated using 5-fold cross-validation is 0.17.}
\label{tab:movie_cnn}
\end{table}

\begin{figure}[t]
  \centering
  \includegraphics[width=0.45\textwidth]{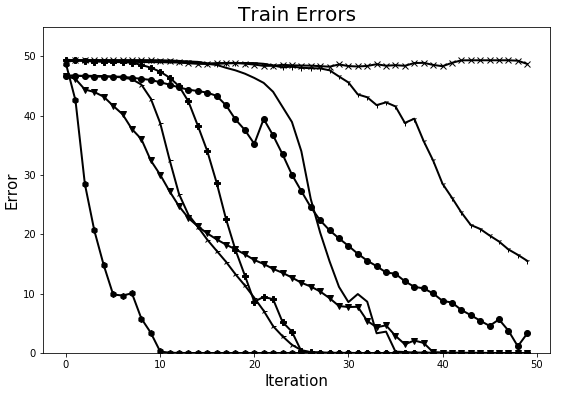}
  \includegraphics[width=0.45\textwidth]{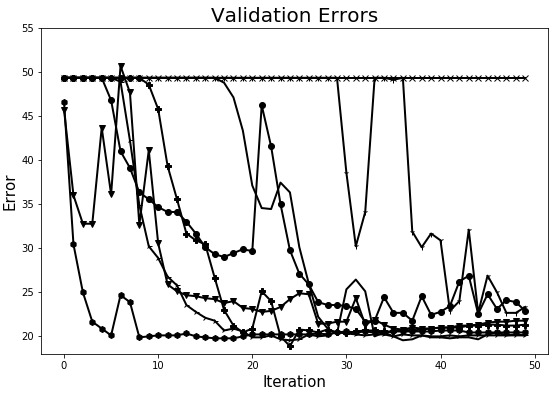}
  \caption{Training and validation error curves of different adaptive models on Movie review data.}
  \label{fig:traincost}
\end{figure}
\label{sect:movie}

"Accordingly, our empirical results show that using adaptive Gumbel, as the activation function, in fully-connected layer is a good choice. Moreover, adaptive ReLu could work excellent when it is applied in convolutional layers. 
A comparison between the best results of our experiments (reported from 5-fold crossvalidation) and the state-of-the-art results by \citep[][]{kim2014convolutional} on the same version of Movie
Review data shows that our adaptive activation functions
perform fairly good and are match on Movie Review data in terms of accuracy. Applying those adaptive activation functions with fine-tuning the hyper parameters may result in further improvement."

\section{Application}
\label{sect:url}
The sequence of URL clicks are gathered to study user intentions by an international tech company. The dataset is private and is extracted from a survey, while anonymous visitors visited a commercial website over a period of three months and their action is recorded at the end of the session.  In those surveys, visitors are typically asked to tell their purpose of visit from the ``browse-search product", ``complete transaction-purchase", ``get order-technical support" or ``other", so treated as a four-class problem.  The study aims at exploring the relationship between user behavioral data (which includes URL sequences as features). The stated purpose of the visit provided in the survey after the end of sessions.  The data consist of approximately 13500 user sessions and each record is limited to maximum of 50 page visits.  The goal is to construct a model that predicts visitors' intention based on URL sequence they navigated from page to page.

The application on user intention prediction are just about to make their early  steps \citep[][]{liu2015convolutional, vieira2015predicting, korpusik2016recurrent, lo2016understanding, hashemi2016query}. Our approach in this work is to use neural networks in two steps; like in the Movie review data of Section~\ref{sect:movie}, by i) embedding a URL into representative vectors and ii) using  these representations as features to a neural network to predict the user intention. We treat each URL as a sentence, where each word of this sentence is separated by ``/".  A similar approach is used in text classification, sentiment analysis \citep[][]{kim2014convolutional}, semantic parsing \citep[][]{yih2014semantic} and sentence modeling \citep[][]{kalchbrenner2014convolutional, kim2014convolutional}.  Again we applied LeNet5 architecture with adaptive activation.  We report \emph{precision} and \emph{recall}, because different methods compete very closely in terms of accuracy.  

The results summarized in Table \ref{tab:url} show that adaptive Gumbel performs slightly better, while ReLU and adaptive ReLU closely compete with each other. 
\begin{table}[ht!]
\centering \small
\begin{tabular}{|l| c c | c c | c c|}
\hline
Conv layer & \multicolumn{6}{|c|}{Fully-connect layer}\\
      & \multicolumn{2}{ c| }{ReLU}&\multicolumn{2}{ c| }{AReLU} & \multicolumn{2}{c|}{AGumb}\\
      & P & R  & P & R & P & R\\
\hline
ReLU &  { 68.0}& { 68.0}& 68.1& 67.9& 68.0& 67.8\\
AReLU &  68.1 & 67.9&  { 68.0}&  {68.0} & 67.9& 67.9 \\
AGumb & { 68.1} & {68.3}& { 68.2}&  { 68.1}&  {68.1}&  {\bf 68.3}\\
\hline
\end{tabular}
\caption{Running LeNet5 convolutional model with ReLU, adaptive ReLU (AReLU), and adaptive Gumbel (AGumb) activations on URL data. Different activations performed almost identical in terms of accuracy, so only the prescition (P), and the recall (R) are reported.}
\label{tab:url}
\end{table}

\subsection{Conclusion}
We proposed a general method to adapt activation functions by looking at the activation function as a cumulative distribution function. This view is useful to adapt  a bounded activation such as sigmoid or hyperbolic tangent.  It is well known in deep neural networks with bounded activations suffer from \emph{vanishing gradient}. Therefore, a methodology for adapting unbounded activations is as important.  We recommended to decompose ReLU into an unbounded component and a bounded component. Therefore, the  cumulative distribution function idea can be re-used to adapt the bounded counterpart. 

In fully-connected networks adapting activation helps prediction most of the time. According to our experiments adapting activation helps prediction accuracy often, even in complex  architectures. However, adaptive Gumbel mostly outperforms other activations in convolutional architectures.  

"In this paper, first, we design a series of experiments to understand how our proposed activation functions affect on the prediction of the fitted models using a simulated data. The results of this experiment show that using adaptive activation functions in fitting models for prediction approximation is superior compared to standard functions in terms of accuracy regardless of network size and activation functions used in original model.
In the next step, we aim at evaluating the performance of the typical CNN models using our proposed activation functions on image and text data. Accordingly, we design a series of experiments on MNIST as a widely-used image benchmark to understand how accurate the adaptive activation functions in LeNet5 CNN models classify the hand-written digit images compared to standard activation functions. The results obtained from this experiment suggest compared to standard sigmoid, applying adaptive Gumbel in fully-connected layer of the CNN models are recommended. Generally, CNN models using proposed activation functions work excellent in prediction and convergence compared to models that work exclusively with standard functions. Additionally, a series of experiments using CNN models trained on a top of word2vec text data is performed to evaluate the performance of the proposed activation functions in a sentiment analysis application on Movie Review benchmark. Our empirical results imply that using adaptive Gumbel as activation functions in fully-connected layer and adaptive ReLu in convolutional layers are strongly recommended. These observations are consistent with the findings were noticed in experiments on MNIST data. Also, a comparison between our best observations and the state-of-the-art results in (Kim, 2014) indicates that our reported results using adaptive activation functions are match with the results reported in this paper. We believe that applying more fine-tuning hyper parameters and using other complex variants of CNN models accompanied with our proposed activation functions could improve the existing results.
To recap, our empirical experiments on two well-known image and text benchmarks imply that by virtue of using adaptive-activation functions in CNN models, we can improve the performance of the deep networks in terms of accuracy and convergence."

Learning the adaptation parameter is feasible by adding only one equation to back-propagation. Computationally,  letting neurons of a layer choose their own activation function, in this framework, is equivalent to adding a neuron to a layer. This minor extra computation changes the network flexibility considerably, especially in shallow architectures. We focused only on the classic LeNet5 architecture, but there is a potential of exploring this methodology with a wide variety of distribution functions for  portable architectures such as MobileNets \cite{howard2017mobilenets}, ProjectionNets \cite{ravi2017projectionnet}, SqueezeNets \cite{iandola2016squeezenet}, QuickNets \cite{ghosh2017quicknet}, etc. 

\section*{Appendix}
Proof of Theorem~\ref{theo:iden}.
Suppose the Bernoulli distribution
\[
	p(y_i) = \pi_i^{y_i}(1-\pi_i)^{(1-y_i)} \texttt{   , } y_i = \{0, 1\},
\]
where $\pi_i$ is a function of parameters $\omega = (\eta, \alpha)$ where $\eta$ is the linear predictor, and $\alpha$ is the activation shape
\begin{equation}
\pi_i = \sigma_\alpha(\eta)
\label{eq:4.30}
\end{equation}
Therefore, to ensure distinct probability distributions are indexed by $\alpha$ on a continuum of $\alpha>0$, the identifiability of $\sigma_\alpha(x)$ must be studied, i.e. distinct values of parameter $\alpha$ lead to distinct activation functions. Formally, 
\begin{equation}
	\alpha \neq \alpha' \leftrightarrow \exists x\in\real \textrm{~~s.t.~~} \sigma_\alpha(x) \neq \sigma_{\alpha'}(x).
\end{equation}
Equivalently 
\begin{equation}
	\sigma_\alpha(x) = \sigma_{\alpha'}(x) \leftrightarrow \alpha = \alpha'\label{eq:4.28}
\end{equation}
 which falls on $\sigma_\alpha(x)$ identifiability definition \citep[][]{huang2005model}. Take  $\sigma_\alpha(x)$ in equation \eqref{eq:adagumbel} that defines adaptive Gumbel
\begin{eqnarray*}
\sigma_\alpha(x) &=& \sigma_{\alpha'}(x),\\
\{1+\alpha\exp(x)\}^{\frac{1}{\alpha}}&=& \{1+\alpha'\exp(x)\}^{\frac{1}{\alpha'}}.\\
\end{eqnarray*}
Given $1+\alpha \exp(x) >0$ 
\begin{eqnarray*}
	\frac{1}{\alpha}\log\{1+\alpha\exp(x)\} &=& \frac{1}{\alpha^\prime}\log\{1+\alpha^{\prime}\exp(x)\}.  \label{eq:4.32}
\end{eqnarray*}

Let $z = \exp(x)$ and $f_z(\alpha) = \log(1+\alpha z)$ which has derivative of arbitrary order. Take the Taylor expansion of $\log(1+\alpha z)$
\[
	\frac{1}{\alpha}\left\{\sum_{n=0}^{\infty}\frac{f_z^{(n)}(0) \alpha ^n}{n!}\right\} = \frac{1}{\alpha^\prime}\left\{\sum_{n=0}^{\infty}\frac{f_z^{(n)}(0) {\alpha ^\prime}^n}{n!}\right\}
\]
The latter equation holds for all $z$ if and only if  all corresponding polynomial coefficients are equal. Equivalently, 
\begin{equation}
	\alpha = \alpha^\prime. \label{eq:4.35}
\end{equation}
\bibliography{farnoush_ref}
\bibliographystyle{humannat}

\end{document}